# Chaotic Variational Auto encoder-based Adversarial Machine Learning

**Pavan Venkata Sainadh Reddy[1,2], Yelleti Vivek[1], Gopi Pranay[1], Vadlamani Ravi[1,*]**

*[1]Centre for Artificial Intelligence and Machine Learning*
*Institute For Development And Research In Banking Technology(IDRBT),*
*Castle Hills Road #1, Masab Tank, Hyderabad 500076, India*
*[2]School of Computer Science And Information Sciences,*
*University of Hyderabad, Hyderabad 500046, India.*
pavansainadh.daka@gmail.com; yvivek@idrbt.ac.in ; gopipranay@idrbt.ac.in; vravi@idrbt.ac.in

## Abstract

Machine Learning (ML) has become the new contrivance in almost every field. This makes them a target of fraudsters by various adversary attacks, thereby hindering the performance of ML models. Evasion and Data-Poison-based attacks are well acclaimed, especially in finance, healthcare, etc. This motivated us to propose a novel computationally less expensive attack mechanism based on the adversarial sample generation by Variational Auto Encoder (VAE). It is well known that Wavelet Neural Network (WNN) is considered computationally efficient in solving image and audio processing, speech recognition, and time-series forecasting. This paper proposed VAE-Deep-Wavelet Neural Network (VAE-Deep-WNN), where Encoder and Decoder employ WNN networks. Further, we proposed chaotic variants of both VAE with Multi-layer perceptron (MLP) and Deep-WNN and named them C-VAE-MLP and C-VAE-Deep-WNN, respectively. Here, we employed a Logistic map to generate random noise in the latent space. In this paper, we performed VAE-based adversary sample generation and applied it to various problems related to finance and cybersecurity domain-related problems such as loan default, credit card fraud, and churn modelling, etc., We performed both Evasion and Data-Poison attacks on Logistic Regression (LR) and Decision Tree (DT) models. The results indicated that VAE-Deep-WNN outperformed the rest in the majority of the datasets and models. However, its chaotic variant C-VAE-Deep-WNN performed almost similarly to VAE-Deep-WNN in the majority of the datasets.

**Keywords**

WNN, VAE-Deep-WNN, Evasion Attack, Data-Poison Attack, Chaotic Maps

---

[*] Corresponding Author

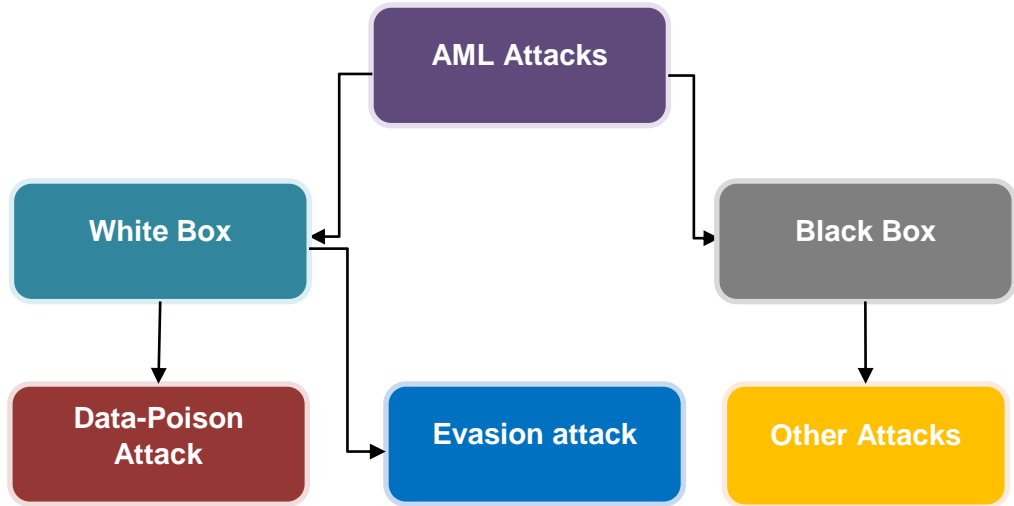

**Figure 1. Taxonomy of AML Attacks**

# 1. Introduction

Nowadays, adversaries are consciously trying to dwindle the Machine Learning (ML) models' performance by using various techniques. Hence, along with gaining accuracy, maintaining the models' robustness gained a lot of attention from practitioners and researchers. A classical example is Email spam filters where the adversary tries to tailor the Email to avoid spam detection. This could depreciate the performance of the model thereby affecting trust and confidence. In such cases, one could employ various defense techniques which could increase the models' robustness. This gave birth to a new field called Adversarial Machine Learning (AML).

**AML** is broadly categorized into two different branches viz., attacks and defenses. Attack techniques (refer to Fig. 1) aim to trick the machine learning models into false training by synthesizing the input. On the other hand, the defenses' primary goal is to make machine learning classifiers aware of these types of attacks. Without limiting to the above classical example, this is very prominent in other fields such as intrusion detection, insider threat detection, and adversary samples to induce bias in the chatbots, etc., This is more prevalent in the fields such as finance, health care, etc., Attacks are categorized into two different types: (i) White box and (ii) Black box attacks respectively. The following assumptions are made in White-box attacks: (i) access to the data, and (ii) the underlying ML model is known and has access to it. However, the black box doesn't have access to either of them. A white box is further divided into data Data-Poison attack and Evasion based on the nature of attacking during the training phase or test phase. To perform data Data-Poison attacker has the access to training data which is used to generate synthetic samples (Chakraborty et al., 2018) and later used to misclassify the model. However, attackers having access to test data are perturbed and utilized to decrease the model performance is called an Evasion attack (Chakraborty et al., 2018).

According to a **Gartner** survey, "*Application leaders must anticipate and prepare to mitigate potential risks of data corruption, model theft, and adversarial samples*"**,** in its first report on adversarial machine learning. Kumar et al. (2020) took a survey which is conducted on 28 industries, ten out of 28 enterprises picked the Data-Poison attack (refer to Fig. 1) that would most affect them and Only 3 industries use secure machine learning which is robust to AML. This shows how AML is associated with and can affect the real world. Owing to its importance and popularity, we focused on the current study's data Data-Poison and Evasion attack. In a recent report by the **Global Risk Institute,** they discussed the attacks on ML in Finance are trading, fraud detection, robo-advising, (Rubstov, 2022) and Natural language Processing where they mentioned replacing a single word in a text sentiment analysis by its synonym can make the model from 99% positive to 100% negative (Morris et al., 2022). Recently, Data-Poison attacks are on the Tay chatbot, which is for communication with users on Twitter and is also designed to retrain with Twitter users' data to get better engagement. In less than 24 hours, Tay learned from normal behavior to unethical behavior. "Microsoft claimed Tay has been attacked by internet trolls and thereby noticed that "the system had insufficient filters and began to feed profane and offensive tweets into Tay's machine learning algorithm" (Centre for long-term CyberSecurity Berkeley[1]).

The original data can be perturbed by various means such as: adding noise, using generative models such as Variational Auto Encoder (VAE), Generative adversarial network (GAN), etc., Among them, generating attack samples from a generative model is quite a popular and effective technique. GAN is relatively highly complex and prone to unstable training(Saxena et al., 2021), on the other hand, VAE is relatively less complex and training is also relatively faster when compared to that of GANs. Further, in the context of AML, VAE is left unexplored. Owing to its advantages and simple-to-use motivated us to propose VAE-based white-box attacks and applied to various financial-related problems such as churn modeling, loan default, credit card fraud detection, and other cyber security problems like distributed denial of service (DDoS) attacks, and intrusion detection systems.

Multi-layer perceptron (MLP) is often used in VAEs' encoder and decoder. Wavelet Neural Network (WNN) is used as an alternative for MLP with lesser parameters and faster training, better functional learning, function approximation, faster conversion rate, and better non-linear function approximation (Venkatesh et al. 2022). However, deeper layers could also be added, called Deep-WNN (Said et al 2016). This motivated us to utilize and propose a novel architecture, VAE-Deep-WNN, where WNN is in both the encoder and decoder. Recently, Chaotic VAE (Gangadhar et al., 2022), is proposed and applied to one class classification (OCC) to perform insurance fraud detection. Taking inspiration from the authors, we proposed Chaotic VAE variants in the context of AML. We designed the corresponding chaotic variants of VAE-MLP and VAE-Deep-WNN and named them C-VAE-MLP and C-VAE-Deep-WNN respectively. All of these models will be discussed in detail in the latter sections. Consequently, all of these models are utilized to perform data Data-Poison attack and Evasion attack respectively, and compared the performance thereof.

---

[1] https://cltc.berkeley.edu/aml/

The major contributions of the current research study are as follows:
- Proposed VAE-Deep-WNN to generate adversary samples and performed Evasion and Data-Poison attacks on Logistic Regression and Decision Tree models.
- Proposed Chaotic inspired variant, C-VAE-Deep-WNN and C-VAE-MLP to perform attacks.
- Further, compared the performance of the above models with VAE-MLP.
- Validated the proposed models in financial and cyber security domain-related problems.

The rest of the paper is organized as follows: In Section 2, we review up-to-date relevant papers. In Section 3, we described the base models VAE, C-VAE, and WNN followed by the proposed methodology. In Section 4, we presented the description of the datasets. Section 5 discusses the results and draws some comparisons between the models. In Section 6, we discuss conclusions and future work on the methodologies.

## 2. Literature Review

Machine learning-based malware analysis techniques are used in various fields like windows malware, android malware for classifying evasive and growing malware threats, however, a small perturbation leads to misclassifying the malicious threats (Aryal et al., 2022). Adversarial attacks are divided into active and passive attacks where the active attacks (Sadeghi et al., 2020), tend to disturb the learning algorithm and corrupt the data which yields bad model performance. Early attacks are performed by modifying only a small portion of training data, however, recent attacks instead focus on accessing all data and modifying all the data at once (Shen et al., 2019, Fowl et al., 2021).

Adversarial examples are malicious samples in which when added a small perturbation to the original samples makes the machine learning models misclassify. Alexey et al. (2017) proposed the Fast gradient signed method (FGSM) and Basic Iterative Method (BIM), to generate adversarial examples on image datasets. Further, they proposed adversarial training for defenses. The image dataset is constructed from a cell phone camera and applied their attacks on the Inception v3 image classification neural network. To the best of our knowledge, Ballet et al. (2019) first illustrated the usage of adversarial examples and adversarial attacks on tabular data like German Credit, Australian Credit, Default Credit Card, and Club Loan datasets using their proposed method LowProFool to generate adversarial samples.

Adversarial examples are generated on Variational AutoEncoder and Generative Adversarial Network (VAE-GAN) (JernejKos et al., 2017), where the attacker/adversary inputs the original data, and the output of that model is taken as the adversarial example. VAE-GAN generates the adversarial examples and is used to make images misclassify. The output of these generative models is nearer to the input data but not the exact data which is the method they are using to perform the adversarial attack. Xiao et al. (2019) proposed the advGAN i.e., Generative adversarial Network for generating the adversarial examples and performed adversarial attacks on semi-white-box (gray-box) and black-box settings for image datasets.

Wang et al. (2021) proposed Man-in-the-Middle Attack against an ML classifier using generative models, here they proposed VAE to generate the adversarial examples primarily using Decoder in a black-box setting against image datasets. Yerlikaya and Bahtiyar (2022) performed a Data-Poisoning attack by injecting adversarial data into training datasets using Random-label flipping and Distance-based label flipping on four datasets namely Breast-cancer, Instagram-spam-filter, Botnet-detection and Android malware on machine learning algorithms for the first time.

Tabassi et al., (2019), defined an Evasion attack, where the attacker tries to solve the constraint optimization problem like gradient descent to find a small input perturbation that causes a large change in the cost function and which results in output misclassification. Chakraborty et al., (2018) illustrated a data modification attack is a white/gray-box attack that instead of an injection of malicious data directly changes the data on the whole itself.

Cartella et al. (2021) illustrated the Zoo attack on the Fraud Detection dataset for the first time and tried making the fraud class the non-fraud class on the XG-Boost algorithm. They generated adversary samples using the Generic Adversarial Algorithm i.e., Harford et al. (2021) proposed the usage of VAE in the generator of the gradient adversarial transformation network (GATN) to enhance the quality of adversarial samples and performed an attack on (1-NN DTW 1- Nearest Neighbour Dynamic Time Wrapping.) and Fully convolution Network (FCN) and performed attacks on the time series datasets. PuVAE (Purifying Variational AutoEncoder) is a method proposed by Hwang et al. (2020) to purify adversarial examples. This method takes the closest projection of the adversarial example as determining it as the purified samples i.e., the PuVAE outputs the nearest estimation of those images and adversarial images get distance measured and whichever gets less distance is taken as the pure image instead of the adversarial image.

Xiang et al. (2021), proposed a Wavelet-VAE structure to reconstruct an input image and generate adversarial examples by modifying the latent code by adding the perturbation. By first training wavelet-VAE for learning the latent distribution of the input image and then fixing the parameters of the models and encoding the target image to obtain latent distribution and performed attacks against 5 state-of-the-art models VGG19, ResNet-152, DenseNet-201, InceptionV3, and Inception-ResNet V2.

Willetts et al. (2021) talked in the paper about improving the robustness of the VAE for adversarial attacks by applying a margin on a VAE's input space within and using disentangled sample generation which is robust to the perturbation. The anomaly detection and online abnormal detection are based on the deep neural network (Xu et al ., 2020) performed a data-poisoning attack on the RNN-based anomaly detection and made the model obsolete by using the general discrete adversarial sample generation using the FGSM method.

Our current research work is different in the following aspects:
- To the best of our knowledge, this is the first-of-its-kind study focused on VAE-based attacks in the financial domain and in the context of tabular datasets.
- Further, we proposed VAE-Deep-WNN for adversary sample generation. However, Xiang et al. (2021) proposed VAE-WNN for image regeneration they didn't perform any sort of adversarial attacks.
- We proposed Chaotic VAE with MLP and Deep-WNN variants and performed Evasion and Data-Poison attacks.

## 3. Proposed Methodology

In this section, we discuss the VAE, and C-VAE variants of MLP and WNN employed for data Data-Poison and Evasion attacks.

### 3.1 Overview Of Techniques

#### 3.1.1 Variational AutoEncoder

It is a deep generative model that uses neural networks to capture the input data distribution proposed by Kingma and Welling, (2013). VAE comprises both an encoder and a decoder. It is regarded as the second generation of the Autoencoder where the aim of the network is not only reducing the dimensions but also learning data distribution and regenerating the data from the latent space. The encoder's aim is to learn the data encoding through reduction to latent space. The decoder's aim is to regenerate the data from that latent space. Then the backpropagation is applied in the form of a loss function to get the network better at the data regeneration nearer to the actual data. The loss function used in the backpropagation comprises the loss between generated dataset and the original dataset and the KL divergenz for a better approximation of the generated dataset to the original dataset refers to Eq. 1.

$$L_{vae} = - D_{KL}[q(z|x)||p(z)] + E_q[\log p(x|z)]. \qquad (1)$$

Where '$D_{KL}$' is kl-divergence(generated data $q(z|x)$.

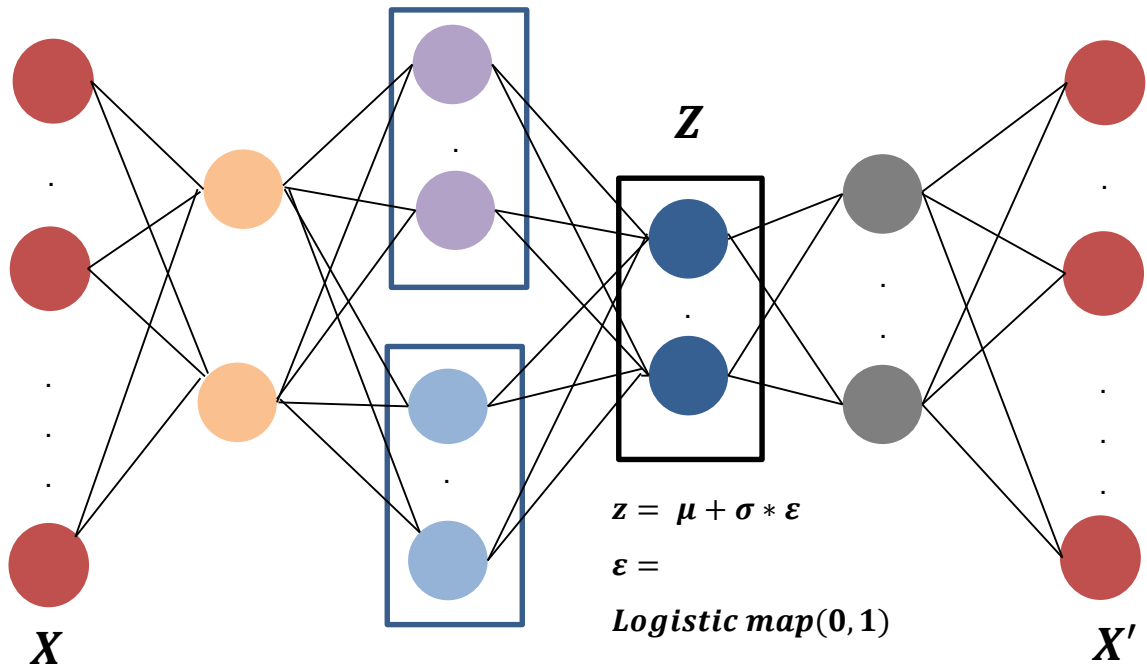

**Figure 2. Chaotic-Variational Autoencoderz**

### 3.1.2 Chaos Theory

Chaotic (Weggins et al., 1963) is defined as "*A dynamical system which depends on small dependence on the initial conditions with a closed invariant set (which consists of more than one orbit) will be Chaotic*". This says that chaotic systems are highly affected even with a small change in the initial conditions. This chaotic behavior can be modeled using non-linear dynamic maps called chaotic maps. Mainly chaos has two properties: ergodicity i.e., the system cannot be reduced into two or more subsets and intrinsically stochastic nature.

Recently, Chaotic maps are being used effectively in Feature subset selection problem (Vivek et al., 2022), handling imbalance (Kate et al., 2022), in handling the one-class classification (OCC) (Gangadhar et al., 2022). In our paper, we utilized a Logistic map which is one of the popular Chaotic maps, and solved various problems (Vivek et al., 2022, Kate et al., 2022, Gangadhar et al., 2022).

**Logistic map:** It is a discrete map of a polynomial equation (refer to Eq. 2) of degree 2. It uses the previous values to generate the next numbers. The nature of the logistic map is decided by $\lambda$. The logistic map exhibits chaotic behavior when $\lambda$ [3.56, 4]. In the current settings, we used '$\lambda$' as 4.

$$x_{t+1} = \lambda * \left( x_t * (1 - x_t) \right) \tag{2}$$

## 3.2 Deep Wavelet Neural Network

Wavelet Neural Networks (WNN), hidden units called wavelons, are primarily based on wave processing having parameters that are used in wavelet analysis. WNN has more success rate than Neural Networks in the fields like Wave Synthesis, Speech Processing, etc. Further, Wavelet neural network is also better at function learning, and the convergence rate is also faster (Zhang et al., 1995). Formally, there are two different operations that are extensively used in signal processing (i) Translation and (ii) Dilation.

**Table 1. Wavelet Functions and properties**

| Sl.No | Wavelet Functions | Equation |
|---|---|---|
| 1 | Morelet (Chauhan et al., 2019) | $f(x) = cos(1.75x) * \exp\left(-\frac{x^2}{2}\right)$ |
| 2 | Gaussian (Chauhan et al., 2019) | $f(x) = \exp(-x^2)$ |
| 3 | Mexican-hat (Wang et al., 2013) | $f(x) = \frac{2}{\sqrt{3}} * \pi^{-0.25} * (1 - x^2) * \exp\left(-\frac{x^2}{2}\right)$ |
| 4 | Shannon (Wang et al., 2013) | $\frac{\sin \pi(x - 0.5) - \sin 2\pi(x - 0.5)}{\pi(x - 0.5)}$ |
| 5 | GGW (Gilbert, Gutierrez, Wang) (Wang et al., 2013) | $\sin 3x + \sin(0.3x) + \sin(0.03x)$ |

In general, WNN consists of three layers: one input layer and one hidden layer, and one output layer. But, researchers (Said et al., 2016) enabled to use of multiple hidden layers and proposed Deep WNN. They applied it for image classification, which yielded better results than single-layered architecture. When multiple hidden layers are used, then it is treated as Deep-WNN, which enables it to capture the underlying complex patterns in a better way than single-layered WNN. WNN uses backpropagation to update its weights.

Unlike, MLP, and WNN, there are separate activation functions (refer to Table 1) called Wavelet Functions, such as Morlet Function, Gauss Function, Mexican Hat Function, etc. This helps to capture underlying non-linear complex patterns.

## 3.3 Proposed VAE-Deep-WNN

The proposed VAE-Deep-WNN also has three different components as VAE encoder, decoder, and latent space, this proposed architecture is different from the VAE in both encoder and decoder parts i.e., the network of both encoder and decoder has to change its internal neural nets with wavelet layers with wavelons and different sets of activation functions.

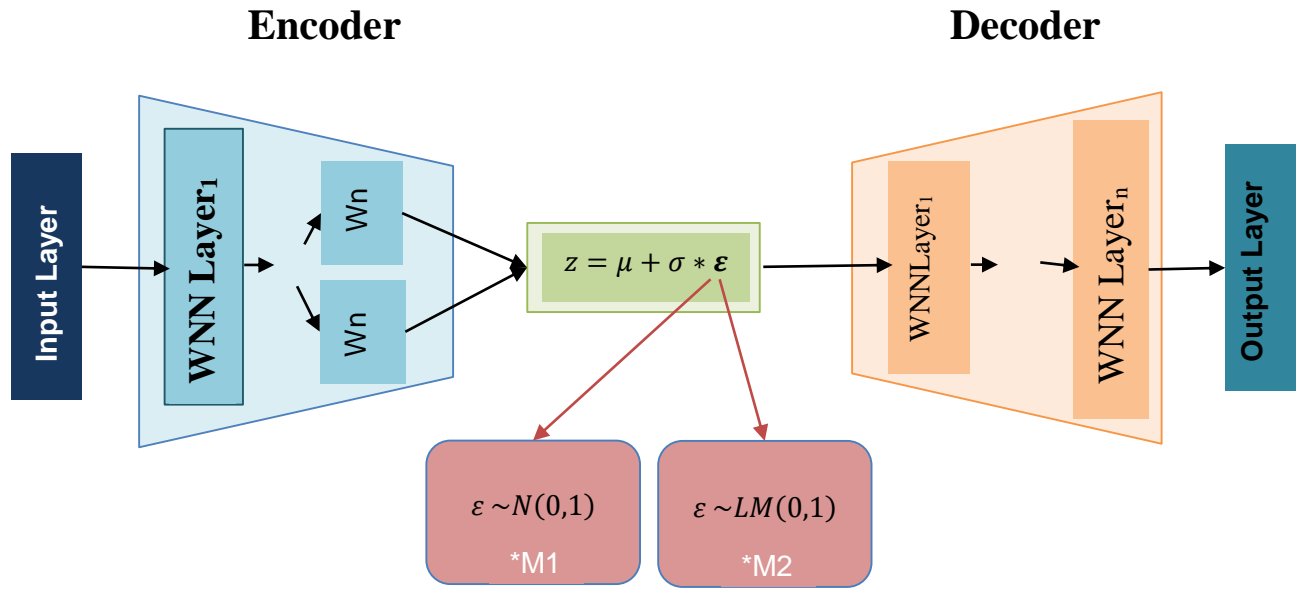

*M1 -> VAE-MLP,VAE-Deep-WNN and *M2 -> C-VAE-MLP,C-VAE-Deep-WNN

N(0,1) ~ Normal Distribution(0,1) and LM(0,1) ~ Logistic-Map(0,1)

**Figure 3. VAE-Deep-WNN / C-VAE-Deep-WNN Architecture**

### 3.3.1 Encoder

In the encoder for each data point from the input is multiplied by the weight and subtracted by the translation divided by dilation parameters then applies wavelet function, as described in the Deep-WNN we used more than hidden wavelet layers in VAE depicted in Fig.3. It takes training dataset as the input with the size of $n * f$, where n is the number of samples and $f$ is the number of features. Suppose the encoder is built with two hidden layers, this network of the encoder has two additional parameters translation and dilation and the equation is given in Eq. 3.
9

$$encoder(x) = f\left(\frac{\sum_{j=0}^{n1} f\left(w_j \frac{\sum_{i=0}^{n} w_i x_i - b_i}{a_i}\right) - b_j}{a_j}\right) \quad (3)$$

where $f$ is the wavelet function w represents the weights, $b_i$ is the translation and $a_i$ is dilation, $i,j$ represents the size of features

### 3.3.2 Latent Space

As in the encoder, the data point is reduced to latent dimension with $\mu$, $\sigma$ from this ' z' is calculated is given by Eq. 4.

$$z = \mu + \sigma * \varepsilon \quad (4)$$

Where $\mu$ is mean, $\sigma$ is standard deviation, $\varepsilon$ is random noise from the Normal distribution N~(0,1). The above Eq.3. is reparameterization of the data coming from the encoder i.e., $\mu, \sigma$ and '$\varepsilon$' is data generator. For every new generation of normal data distribution we need mean and sigma of that particular normal distribution and a random number generator.

### 3.3.3. Decoder

Now, from the latent space vector 'z' the decoder layer reproduces the data using the same number of hidden layers and wavelons. The layers start in the decoder is a mirage of the encoder in reverse, is given in Eq. 5.

$$decoder(z) = f\left(\frac{\sum_{i=0}^{n} f\left(w_i \frac{\sum_{j=0}^{n1} w_j z_j - b_j}{a_j}\right) - b_i}{a_i}\right) \quad (5)$$

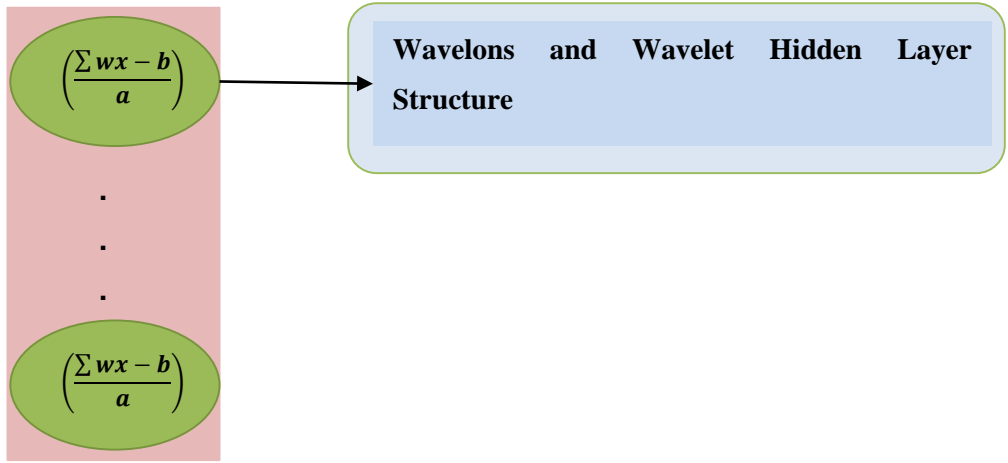

**Figure 4. Wavelet Neural Network hidden layer representation**

## 3.4 Proposed Chaotic-VAE-Deep-WNN

In VAE, there is the flexibility that we can use different distributions in the epsilon part in the reparameterization of the network, based on our need, it is also discussed in Kingma and Welling (2013). The major difference VAE-Deep-WNN and C-VAE-Deep-WNN is the introduction of chaotic maps in the latent space. The Reparameterization part of the VAE model is where the changes take place with respect to the VAE-WNN. The '$\varepsilon$' which is multiplied by '$\sigma$' is generally the standard normal distribution ~N(0,1) due to the fact the input data is normalized between (0,1).

### 3.4.1 Encoder

The encoder is same as the VAE-Deep-WNN that given input is multiplied by weights and subtracted by translation and divided by the dilation parameter (refer to Eq. 3).

### 3.4.2 Decoder

The decoder is also same as the VAE-Deep-WNN, here from calculated 'z' this part tries to reproduce the data from the latent distribution (refer to Eq. 4).

### 3.4.3 Latent distribution

This section is where this model is changed from the VAE-Deep-WNN, generally the '$\varepsilon$'(epsilon) from the reparameterization equation (refer to Eq. 5). the noise is the normal distribution between (0,1) because to use it as the generator of the new distribution of the data, as described in (Kingma and Welling 2013) the noise can changed as per requirement and (Gangadhar et al., 2022) used Logistic map to generate chaotic number for '$\varepsilon$' in C-VAE Here the usage Logistic map (0,1) in epsilon to change of the model method of reproduction of the data.

## 3.5 Attack Methodology

In this work we performed two attacks: (i) Evasion attack, (ii) Data-Poison Attack.

### 3.5.1 Evasion Attack Methodology

Assuming that the attacks environment is **white-box** i.e., as described by (Biggio et al., 2017) that to be assumed that adversary had perfect knowledge about the victim's model, for this attack method the adversary can change the test time samples. The whole process is as follows, using stratified random sampling the dataset is divided into two sets named $X_{train}, X_{test}$ with the ratio of 70:30. Then we normalize the $X_{train}$ using MIN-MAX-scaler for scaling of the data to range between (0,1), we used five datasets refer to table 3. Assumed we are in white-box environment and able to access the datasets out of three datasets needed the use of oversampling due to class imbalance using SMOTE. Fore adversarial sample generation $X'$ the generative model needs to be trained, here we trained it using $X_{train}$ refer to **Algorithm 3.** for VAE the process starts as follows, in the case of VAE-MLP and C-VAE, the weights are initialized and in the case of VAE-Deep-WNN and C-VAE-Deep-WNN, along with weight initialization, translation and dilation parameters is also done. All of these are randomly initialized by using normal distribution N~(0,1). For C-VAE and C-VAE-Deep-WNN the seed value for the chaotic map is fixed.

**Algorithm 1.** Pseudocode for Evasion attack using VAE and its variants

*Input:* X: dataset L: number of epochs, lr: Learning Rate, mom: Momentum
*Output:* $AUC_{evasion}$

1. $X_{train}, X_{test}$ ← *Divide the dataset into train and test data*
2. *Normalize the train and test datasets*
3. $X'_{train}$ ← *Apply Oversampling technique in the case of imbalanced datasets. i.e., SMOTE*
4. $ML_{model}$ ← *Get trained Machine Learning model*
   *# ADVERSARIAL SAMPLE GENERATION*
5. $X'$ ← *Generate Data using VAE / C-VAE with MLP or WNN using $X_{test}$ by calling Algorithm 3.*
6. *Use generated . X' to validate . $ML_{model}$*
7. $AUC_{evasion}$ ← *Compute AUC score*
8. *Return $AUC_{evasion}$*

After initialization, the training of the encoder with VAE models described in the Algorithm 3., based on the variant the adding of the noise in the reparameterization part is changed from Random Normal Distribution(0,1) to Logistic map(0,1) based the normal and chaotic variants. From which decoder starts reproducing the data, after each epoch the loss is calculated between regenerated data and the original data (refer to Eq. 1)**.** Thus calculated loss is used for the backpropagation, if we use regular neural network only weights gets updated with respect to the learning rate/momentum and optimizer used refer to table 3., but in the wavelet neural network there are two additional parameters i.e., translation and dilation also gets updated rest of the process is same as the regular neural network's backpropagation and This process continues until the epochs complete.

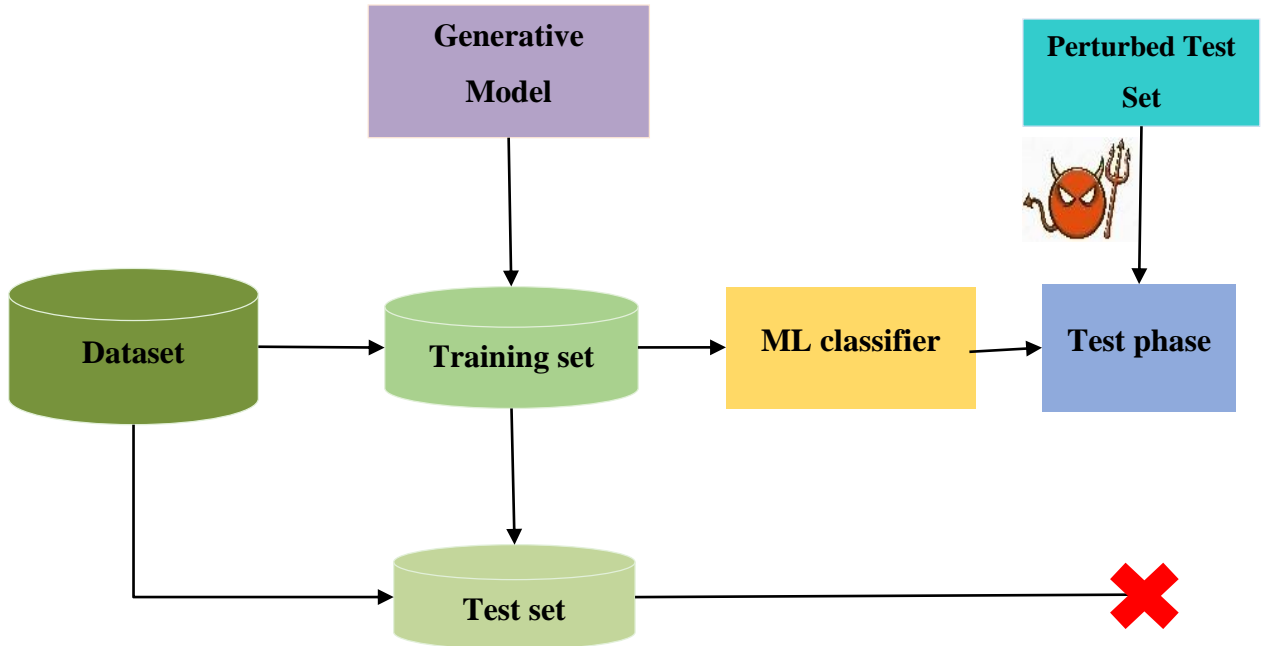

**Figure 5. Schematic diagram of Evasion Attack**

Since the generative model is trained now we need to feed to the training set/test set based on the attack type and the trained generative model gives the adversarial sample $X'$. In this attack test set is given as the input to get the adversarial test-set. The victim machine learning classifier is considered since we are using binary classification datasets refer to table 2., now the victim's model is trained using original training dataset and validated using the adversarial test set generated using VAE and its variants inplace of the original test set then calculation of AUC using the formula refer to Eq. 6. And got the AUC score of the attack's performance and the victim's model original performance needed to be checked using the original test set for validation and calculating the AUC score using formula refer to Eq .6. Refer to schematic diagram Figure. 5. For the representation of the whole process.

### 3.5.2 Data-Poison Attack Method

This attack is also assumed to be white-box and the adversary has the perfect knowledge of the data and data distribution of the victim's model, the whole process up to adversarial sample generation is the same as the Evasion attack methodology, the parameters for VAE-MLP, C-VAE like weights are initialized using a random normal distribution(0,1) and the initial seed for the chaotic variant needs to be fixed and for the VAE-Deep-WNN and C-VAE-Deep-WNN the additional parameters translation and dilation needs to initialized using random normal distribution (0,1).

**Algorithm 2.** Pseudocode for Data-Poison attack using VAE and its variants

*Input:* X: dataset, L: number of epochs, lr: Learning Rate, mom: Momentum
*Output:* $AUC_{poison}$
1. $X_{train}, X_{test}$ ← *Divide the dataset into train and test data*
2. *Normalize the train and test datasets*
3. $X'_{train}$ ← *Apply Oversampling technique in the case of imbalanced datasets. i.e., SMOTE*
4. $ML_{model}$ ← *Get Machine Learning model*
5. $X'$ ← *Generate Data using VAE / C-VAE with MLP or WNN using $X_{train}$ by calling Algorithm 3.)*
6. *Train derived . $ML_{model}$ using $X'$*
7. *Validate the performance of $ML_{model}$ using $X_{test}$*
8. $AUC_{poison}$ ← *Compute AUC score*

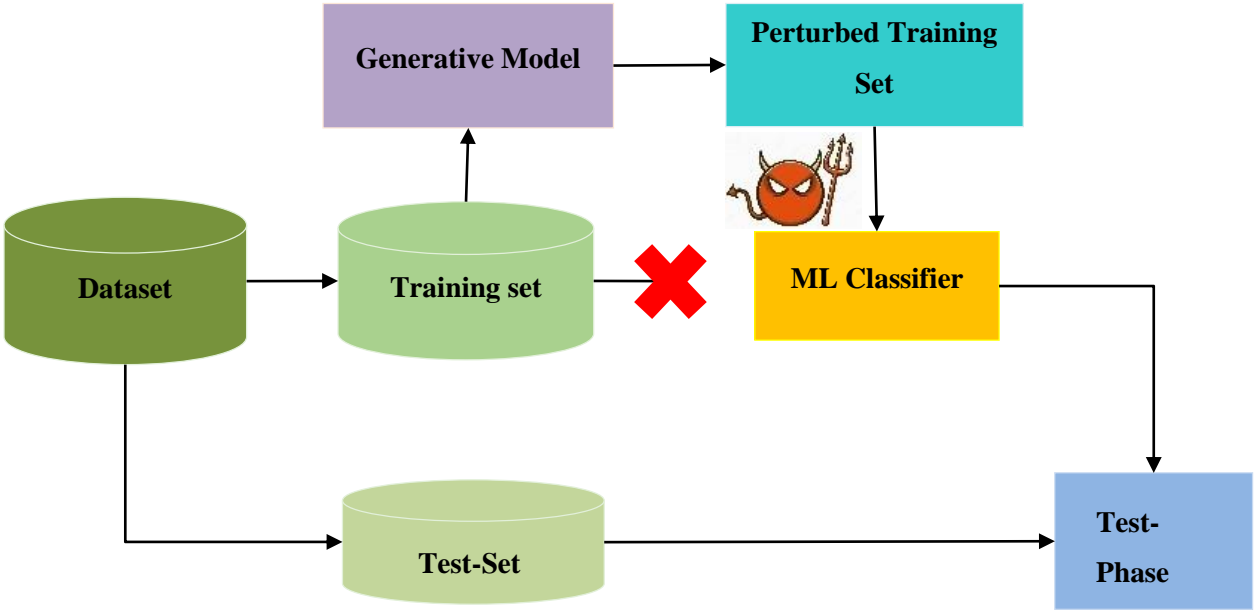

**Figure 6. Schematic diagram of Data-Poison Attack**

Oversampling using SMOTE is applied based on the dataset's imbalanceness. From which regeneration of data starts, and loss refer to Eq. 1. is calculated after each epoch. Based on the loss the backpropagation with respect to weights in the VAE and C-VAE and weights, translation, and dilation for the VAE-Deep-WNN and C-VAE-Deep-WNN.

For the adversarial sample generation same as the above section the generative model VAE and its variants is trained using the training set and this trained VAE model given the original $X_{train}$ as the input and adversarial samples is generated as the output. Since we are using the binary classification datasets the Victim model is considered as the classification based model the victim model is trained on the original dataset and validated on the original test set, for which the AUC score is calculated using the formula refer to Eq . 6. , and these scores considered as before attack scores and the model is retrained using modified adversarial training set $X'$ by adversary, given as the training dataset to the victim model and validated on the original test set, for which the AUC score is calculated using the formula refer to Eq . 6. , for four variants is collected is, and these values are considered as After attack's AUC, the whole process represented using the schematic diagram refer to Fig. 6.

**Algorithm 3.** Pseudocode of the adversarial sample generation using VAE and its variants

**Input:** $X_{train}$: training dataset, $X_{test}$: test dataset, L: number of epochs, lr: Learning Rate, mom: Momentum

**Output:** $M_{[\phi,\theta]}$: trained model, Attack_$X'$: perturbed data

1. $M_{[\phi,\theta]} \leftarrow \emptyset$     // initialize the Generative model (C-VAE, VAE);
2. **for** k is 1 **to** L **do**
    a. x ← Randomly obtain the dataset from $X_{train}$
    b. $\epsilon$ ← Train the Encoder model
    c. $\rho$ ← Generate random noise by using Logistic Map
    d. $\zeta$ ← Train the Decoder model by using chaotic random noise and Z
    e. Calculate Loss (refer to Eq. 2)
    f. Perform Backpropagation
    // update model $M_{[\phi,\theta]}$ parameters
3. **end for**
4. $X'_{test}$ ← Test the trained model, $M_{[\phi,\theta]}$ performance on $X_{test}$
5. $X'$ ← Generate the adversary sample dataset using $M_{[\phi,\theta]}$ and $X_{train}$ or $X_{test}$ depending on the type of attack.

# 4. Dataset Description

In the current study, we evaluated the performance of the proposed models with VAE-MLP on five datasets associated with the financial and cyber security domains. All the datasets are presented in Table 1. All of them are binary classification datasets, and the corresponding disparities are also presented.

**Table 2. Datasets Description**

| Datasets Name | No of classes | No of Samples | No of Features | Downloaded from |
|---|---|---|---|---|
| **Bank-Churn** | 2 | 10,000 | 12 | Kaggle[2] |
| **Credit-card Default** | 2 | 30,000 | 23 | Kaggle |
| **Loan-Default** | 2 | 30,000 | 25 | Kaggle |
| **CICIDS-2018** | 2 | 2,00,000 | 66 | CIC[3] |
| **CICDDOS-2019-DNS** | 2 | 50,00,000 | 88 | CIC |

Further, bank churn and loan default datasets are balanced and the rest of the datasets are highly imbalanced. We incorporated Synthetic Minority Oversampling Technique (SMOTE) to handle the imbalance problem. The dataset is divided into training and test datasets in 70:30 proportion with a stratified random sampling method.

**Table 3**. Hyperparameters for all Techniques

| Model | Hyperparameters |
|---|---|
| **VAE-MLP / C-VAE-MLP** | Epochs:[100,200,500,1000,1400,1500,1800]<br>#Number of hidden layers :[1,2,3,4]<br>Learning rate :[0.0001,0.001,0.01,0.05]<br>Momentum :[0.001,0.01]<br>Activations:['relu,' Tanh']<br>Optimizers :[' Adam',' Adagrad',' SGD']<br>Latent-Dimensions :[2,3,4,8] |
| **VAE-Deep-WNN / C-VAE-Deep-WNN** | Epochs:[100,200,500,1000,1400,1500,1800]<br>#Number of hidden layers :[1,2,3,4]<br>Learning rate :[0.0001,0.001,0.01,0.05]<br>Momentum :[0.001,0.01]<br>Wavelet-Functions:['Morlet',' Gaussian',' Mexican-hat',' Shannon','ggw']<br>Optimizers :[' Adam',' Adagrad',' SGD']<br>Latent-dimensions:[2,3,4,6,8] |

# 5. Results and Discussion

In this section, we will thoroughly investigate the performance of the VAE and C-VAE variants with MLP and WNN in Evasion and Data-Poison attacks. The hyperparameters with which the models are trained are presented in Table 3. First, we will discuss the Evasion attack followed by the Data-Poison attack. As discussed earlier, we employed stratified random sampling and maintained the proportion as 70:30 for the training and test dataset. And Normalized the training set and test set with the use of Min-max-scaler for re-scaling the data between (0,1) and SMOTE is also employed wherever applicable. In our study, we performed analysis on two different ML techniques viz., Logistic Regression (LR), and Decision Tree (DT) respectively. These two models are chosen because these models are chosen to be effective and efficient while handling tabular datasets (vivek et al., 2022). It is a known fact that the Area under the receiver operator characteristic curve (AUC) is robust to imbalance. Hence, we considered AUC as the metric in analyzing the performance of the models.

---

[2] https://www.kaggle.com/datasets ;
[3] https://www.unb.ca/cic/datasets/index.html ;

## 5.1 Classification Models

### 5.1.1 Logistic regression

Logistic Regression (LR), employs a supervised classification technique for the binary classification task. LR is also known as the logit model, since it estimates the probability of an event occurring such as being fraudulent or legitimate, churned out or non-churn customer, etc. LR assumes that all the features are independent of each other. Even though this is a simple model yet most of the time it is proven to be effective when applied to tabular datasets.

### 5.1.2 Decision Tree

DT (Breiman et al., 1984) is one of the most powerful techniques used to solve various machine-learning tasks such as classification, and regression. DT generates a tree structure based on the training data comprising internal and leaf nodes. Each internal node serves as the classification rule, and each leaf node denotes a class label. The following are a few of the advantages of DT: handling missing values, automated feature selection, and handling both categorical and numerical features. The splitting of the root nodes is based on different criteria such as Gini, entropy etc., and DT generates a set of rules which makes it easily interpretable. Hence, DT is considered as the white box model. This model is more popular to be described in detail.

## 5.2 Evaluation Metric

In the current work, AUC is chosen to be the evaluation metric. AUC is proven to be a robust measure while handling unbalanced datasets and is an average of specificity and sensitivity. The mathematical representation of AUC is given in Eq. (6). Sensitivity (refer to Eq. (7)) is the ratio of the positive samples that are truly predicted to be positive to all the positive samples. Specificity (refer to Eq. (8)) is the ratio of the negative samples that are truly predicted to be negative to all the negative samples.

$$AUC = \frac{(Sensitivity + Specificity)}{2} \quad (6)$$

Where,

$$Sensitivity = \frac{TP}{TP + FN} \quad (7)$$

and

$$Specificity = \frac{TN}{TN + FP} \quad (8)$$

Where TP is a true positive, FN is a false negative, TN is a true negative, and FP is a false positive.

## 5.3 Evasion Attack

Now, we will discuss the AUC attained by various ML models after performing an Evasion attack by using various generative models. The results are presented in Table 4. As we know, an Evasion attack is performed during the test phase. Hence, by using Algorithm 1, we generated Evasion samples by using test data. Thus generated adversary samples are used to attack LR and DT, respectively. The corresponding results are presented in Table 4. It is desired that post-attack, the AUC should get decreased. The higher the decrease in AUC, the better the generative model can generate the Evasion samples.

**Table 4. AUC attained by various models after performing an Evasion attack**

| Dataset | Model | Before Attack | After Attack | | | |
| --- | --- | --- | --- | --- | --- | --- |
| | | | VAE-MLP | VAE-Deep-WNN | C-VAE-MLP | C-VAE-Deep-WNN |
| Bank-Churn | LR | 0.5670 | 0.5071 | **0.4981** | 0.5076 | 0.5049 |
| | DT | 0.6742 | 0.6016 | **0.4764** | 0.5529 | 0.4862 |
| Loan-Default | LR | 0.5996 | **0.5003** | 0.505 | 0.5747 | 0.5017 |
| | DT | 0.665 | 0.493 | **0.456** | 0.5687 | 0.4657 |
| Credit-Card | LR | 0.88 | 0.545 | **0.510** | 0.5168 | **0.510** |
| | DT | 0.9002 | 0.5223 | **0.5006** | 0.4997 | 0.501 |
| CICIDS-2018 | LR | 0.975 | 0.582 | **0.4993** | 0.5798 | 0.5131 |
| | DT | 0.9832 | 0.5273 | **0.4795** | 0.49123 | 0.4877 |
| DDOS-DNS-2019 | LR | 0.805 | 0.5351 | 0.5223 | 0.5465 | **0.4957** |
| | DT | 1.0 | 0.5245 | **0.4977** | 0.556 | **0.4997** |

The results indicate that all the generative models can affect the model performance with the generated dataset. Among the models, WNN variants outperformed the MLP variants in all of the datasets concerning all of the models. Especially, VAE-Deep-WNN affected the AUC most number of times than its corresponding C-VAE-Deep-WNN. Interestingly, the AUC after the attack is similar for both VAE-WNN and C-VAE-Deep-WNN in most of the cases, yet the former outperformed the latter in the majority of the datasets. However, in the loan default dataset, VAE-MLP performed the best, yet the difference between the next best, i.e., C-VAE-Deep-WNN, is minimal (0.1%). Among all of the VAE variants, C-VAE-MLP performed the worst.

## 5.4 Data-Poison Attack

Now, we will discuss the AUC attained by various ML models after performing a Data-Poison attack using various generative models. The results are presented in Table 5. As we know, a Data-Poison attack is performed during the training phase. Hence, by using Algorithm 2, we generated adversarial samples by using training data. Here also, the generated adversary samples are used to attack LR and DT, and the results are presented in Table 3. Same as earlier, here also, the decrease in AUC is desired after performing a Data-Poison attack. The results indicate that WNN variants outperformed the MLP variants in all of the datasets concerning all of the models. Further, all of the VAE variants affected AUC with a great percentage which empirically proves that the actual attack happened. Likewise previous Evasion attacks, VAE-Deep-WNN affected the AUC most number of times than its corresponding chaotic variant, viz., C-VAE-Deep-WNN. Interestingly, C-VAE-Deep-WNN performed the worst in DDOS-DNS 2019 dataset, where the difference in performance concerning VAE-Deep-WNN is 10%. Except in this case, in all the other datasets with all of the models, VAE-Deep-WNN and C-VAE-Deep-WNN performed almost similarly.

**Table 5. AUC attained by various models after performing a Data-Poison attack**

| Dataset | Model | Before Attack | After Attack | | | |
|---|---|---|---|---|---|---|
| | | | VAE-MLP | VAE-Deep-WNN | C-VAE-MLP | C-VAE-Deep-WNN |
| Bank-Churn | LR | 0.567 | 0.5018 | **0.4981** | 0.5212 | 0.5 |
| | DT | 0.6742 | 0.5735 | 0.513 | 0.4892 | **0.4539** |
| Loan-Default | LR | 0.5996 | 0.5634 | 0.505 | 0.5750 | **0.5017** |
| | DT | 0.665 | 0.4926 | **0.456** | 0.5038 | 0.4931 |
| Credit-Card | LR | 0.88 | **0.50109** | 0.5102 | 0.5852 | 0.53363 |
| | DT | 0.9002 | **0.492** | 0.5006 | 0.5214 | 0.5353 |
| CICIDS-2018 | LR | 0.975 | 0.5840 | **0.4993** | 0.5798 | 0.5002 |
| | DT | 0.9832 | 0.5321 | **0.4795** | 0.5021 | 0.4862 |
| DDOS-DNS-2019 | LR | 0.805 | 0.625 | 0.5223 | 0.6104 | **0.5102** |
| | DT | 1.0 | 0.6341 | **0.4977** | 0.6231 | 0.5987 |

Whereas in the credit-card fraud dataset, VAE-MLP performed the best concerning both LR and DT, the difference between the next best, i.e., C-VAE-Deep-WNN, is very minimal (0.1%) in either of the cases. Among all of the VAE variants, C-VAE-MLP performed the worst. It is observed that the Bank-Churn and Loan-Default the AUC drop is almost around 20% from before and after attack , if we observe the rest of three datasets the percentage drop is almost nearly around 50% which is making the machine learning model from correct prediction to making the model falsely predicting one class to other.

As discussed in the literature survey before that, the WNN is better at function learning and approximation(Zhang et al., 1995) and Harfold et al. (2022) also illustrated that inducing VAE in their adversarial sample generator enhanced the adversarial sample quality; in this current work, the results also indicated the same strengthening the fact that VAE-WNN combined can produce the quality adversarial samples. Given the Chaotic maps-based models recently shown by (Vivek et al., 2022)(Kate et al., 2022) better performed, we also implemented the same with VAE-Deep-WNN for possible better enhancement of the adversarial sample generation though it is shown not so better results than the VAE-Deep-WNN, out of total four variants of VAE is used to generate adversarial sample C-VAE-Deep-WNN outperformed the VAE, C-VAE-MLP shown it is better at the generation of the adversarial samples than these two variants. In the experimentation stage, we used numerous hyperparameters related to wavelets and VAE refer to table 3. The usage of different wavelet functions changed adversarial sample quality and attacks's performance rapidly, out of all wavelet functions overall the morlet wavelet function performed better irrespective of datasets. Due to the wavelet function's no symmetry (Wang et al., 2013) behaviour is the primary reason of the better performance in the attacks.

## 6. Conclusion & Future work

This is a first-of-its-kind method, where we utilized VAE for adversarial sample generation and performed Evasion and Data-Poison attack on financial and cybersecurity domain. We further proposed VAE-Deep-WNN and chaotic based C-VAE-Deep-WNN, for the adversarial sample generation. Among all of the generative models, VAE-Deep-WNN outperformed the rest the majority of the time. However, its chaotic variant C-VAE-Deep-WNN performed almost similar or better than VAE-most of the datasets.In future, we will perform the same VAE based attacks on time series, images, and regression problems. Further, we can perform the proposed attacks on other machine learning models such as Multi-layer perceptron(MLP), Boosting and Bagging Models.